\documentclass[conference]{IEEEtran}
\IEEEoverridecommandlockouts
\usepackage{cite}
\usepackage{amsmath,amssymb,amsfonts}
\usepackage{algorithmic}
\usepackage{graphicx}
\usepackage{textcomp}
\usepackage{xcolor}
\usepackage{booktabs}
\usepackage{multirow}
\usepackage{siunitx}
\usepackage{tikz}
\usetikzlibrary{positioning,arrows.meta,fit,backgrounds,calc}
\def\BibTeX{{\rm B\kern-.05em{\sc i\kern-.025em b}\kern-.08em T\kern-.1667em\lower.7ex\hbox{E}\kern-.125emX}}

\newcommand{\dgt}{\mathbf{D}_{\mathrm{gt}}}
\newcommand{\dhat}{\hat{\mathbf{D}}}
\newcommand{\otask}{\mathbf{o}_{\mathrm{task}}}
\newcommand{\oaug}{\mathbf{o}_{\mathrm{aug}}}
\newcommand{\methodname}{OBS}
\newcommand{\ci}[1]{\textcolor{gray}{\tiny~(\ensuremath{\pm\num{#1}})}}

\begin{document}

\title{Privileged Critic Training Enables Sensor-Free Thruster Fault Adaptation in End-to-End RL}

\author{Ricard M. Castan$^1$, Miguel A. Olivarez-Mendez$^1$%
\thanks{$^1$University of Luxembourg}%
\thanks{$^2$https://github.com/snt-spacer/RAFT.git}
}

\maketitle

\begin{abstract}
Fault-tolerant navigation for thruster-actuated robots requires online adaptation to failures that are neither binary nor fully observable: thrusters may degrade continuously, fail dead, or jam stuck-open. Classical fault detection pipelines require dedicated sensors unavailable at deployment; oracle controllers that observe the true failure state are equally impractical. We show that privileged critic training is sufficient for sensor-free fault adaptation: giving the PPO value function access to the true degradation state $\dgt$ during training, while the actor receives only standard task observations, shapes a policy that compensates for failures at deployment without any dedicated fault sensing. We propose \textbf{RAFT} (Recurrent Asymmetric Fault Tolerant), a policy with recurrent memory trained with a privileged asymmetric critic. Evaluated on a floating-platform robot (8 thrusters, 1 reaction wheel) under up to four simultaneous thruster failures, RAFT achieves \textbf{70.2\,\%} success at four concurrent failures, closing \textbf{84\,\%} of the gap from a failure-naive baseline (4.8\,\%) to an oracle policy that sees the full degradation state at deployment (82.4\,\%). All code, checkpoints, and data are open-source$^2$.
\end{abstract}

\section{Introduction}
\label{sec:intro}

Robots that rely on thrusters for force generation, underwater vehicles, aerial platforms, or free-floating space robots, must operate reliably despite hardware degradation. Real failures rarely present as clean binary switches: a propeller may lose thrust gradually due to biofouling, a valve may jam partially open, or an actuator may fail dead mid-task. Robust operation thus requires online detection and compensation of continuous, multi-mode degradation acting simultaneously on multiple actuators.

Classical fault-tolerant control handles this via an explicit fault-detection and isolation (FDI) module followed by a switched recovery controller~\cite{jiang2006fault,zhang2008bibliographical}. This pipeline is brittle: each fault mode requires a dedicated FDI signature, and combinations of faults quickly overwhelm lookup-table approaches. Data-driven FDI methods~\cite{miljkovic2011fault} relax the need for hand-crafted signatures but still require a second-stage compensator designed for each identified fault.

End-to-end reinforcement learning sidesteps the modular pipeline by training a PPO~\cite{schulman2017proximal} policy directly under failure distributions~\cite{tobin2017domain}. The two most common baselines are an oracle policy (Oracle) that injects the true degradation vector $\dgt$ directly into the actor (an upper bound requiring perfect fault sensing), and a vanilla policy (VAN) trained in failure-free environments. Neither is deployment-ready. We ask a different question: if the value function can observe $\dgt$ during training, does the actor still need any fault information at deployment? In the asymmetric actor-critic framework~\cite{pinto2017asymmetric}, the PPO critic receives privileged access to $\dgt$ while the actor sees only standard task observations. We show empirically that this critic-side privilege is the primary mechanism for sensor-free fault tolerance.

\textbf{Contributions.} (i) \textbf{RAFT}, a privileged-asymmetric-critic GRU policy achieving 70.2\,\% SR at four simultaneous failures (84\,\% of the VAN$\to$oracle gap) without any fault sensors at deployment; (ii) an ablation study (E6) isolating the contribution of critic privilege from actor architecture, where a memory-less feedforward actor with the same critic (VAN-MLP-AC) already achieves 79\,\% gap closure, confirming the critic as the primary mechanism while compact recurrence adds a further 3.8\,pp; (iii) the \textbf{Observer} architecture as an optional interpretability extension with a three-mode failure scheduler, and a quantified tradeoff showing OBS-MSE produces a human-readable fault estimate at a cost of 11\,pp SR relative to RAFT; (iv) a qualitative discussion of where fault tolerance is encoded (policy weights vs.\ runtime state estimation) and the per-mode observability of $\dgt$ from passive motion history.

\section{Related Work}
\label{sec:related}

\textbf{Fault-tolerant control.} Surveys of actuator fault-tolerant control~\cite{jiang2006fault, zhang2008bibliographical} categorize the field into passive approaches (fixed robust controllers) and active approaches (real-time fault detection and isolation followed by controller reconfiguration). Active approaches require an FDI module that reliably identifies fault type and magnitude, which is difficult for simultaneous or continuously degrading failures. Our work sidesteps the FDI module entirely: the asymmetric critic encodes fault-compensating behavior directly into the actor weights during training.

\textbf{Asymmetric actor-critic with privileged information.} The asymmetric actor-critic framework~\cite{pinto2017asymmetric} shows that a critic with privileged simulation state can guide a deployment-ready actor trained on realistic observations. This idea has been scaled to massively parallel GPU simulation for legged locomotion~\cite{rudin2022learning}, demonstrating that critic-privileged training yields robust policies across diverse physical conditions. Work in~\cite{ebi2025informed} proves that conditioning the critic on \emph{arbitrary partial} privileged signals still produces unbiased policy gradient estimates, directly justifying RAFT's design of feeding only the degradation vector $\dgt$ to the critic rather than the full environment state. Finite-time convergence analysis in~\cite{lambrechts2025theoretical} shows that asymmetric critics eliminate aliasing error terms arising from partial observability, giving the first formal convergence guarantee for this training paradigm. Polynomial sample-complexity bounds under a filter stability condition are established in~\cite{cai2024provable}, which also shows that two-stage expert distillation provably fails in the general case, motivating the single-phase critic-privilege design over teacher-student pipelines.

\textbf{History-conditioned adaptation.} RMA~\cite{kumar2021rma} trains a privileged extrinsics encoder in simulation, then distils it into a history-conditioned adaptation module via behaviour cloning. A teacher-student framework in~\cite{lee2020learning} similarly distils a policy trained with privileged terrain information into one operating solely on proprioceptive history. RMA has been extended to robotic manipulation~\cite{liang2024rapid}, demonstrating that hidden physical variables such as object mass and friction can be inferred from action and proprioceptive history without direct sensor access, establishing the generality of privileged-to-history distillation beyond locomotion. Our Observer architecture follows a similar spirit but trains history encoder and policy jointly end-to-end, without a separate distillation phase. LSTM~\cite{hochreiter1997long} and GRU~\cite{cho2014learning} policies can in principle detect failures implicitly through hidden state accumulation. Recurrent model-free RL has been shown~\cite{ni2022recurrent} to match specialized POMDP algorithms across diverse benchmarks, establishing recurrent architectures as strong baselines. Our E5 ablation tests this directly in the fault-tolerance setting: four GRU/LSTM variants without privileged critic access all perform at or below VAN-MLP (best: GRU-64, 4.0\,\% at $k{=}4$). E6 shows that restoring critic privilege to a memory-less actor recovers 66.4\,\%, confirming that critic privilege, not recurrence or an explicit history buffer, is the primary mechanism.

\textbf{Learning-based fault tolerance.} Domain randomization over actuator effectiveness~\cite{tobin2017domain} trains policies robust to sampled failure parameters at training time, closely mirroring our VAN baseline; we show that without privileged critic access, such policies degrade rapidly as failure count grows. Recent work~\cite{lagattu2025control} demonstrates model-free DRL for thruster force reallocation on an AUV experiencing degradation or loss, without an explicit FDI system, showing the viability of end-to-end learning for continuous actuator faults in underwater vehicles. Our setting differs in that we study simultaneous multi-thruster failures across three distinct failure modes on a planar free-floating platform, and explicitly isolate the contribution of asymmetric critic training.

\textbf{Underwater and space robotics.} Fault-tolerant control for AUVs has been studied in classical~\cite{fierro1998control} and model-predictive frameworks~\cite{shen2018model}. More recent work~\cite{hamamatsu2025cross} demonstrates an RL-based fault-tolerant surfacing controller for underwater robots that operates without any explicit fault identification and transfers across three different AUV designs, achieving 85.7\,\% real-world success and establishing the feasibility of sensor-free RL on free-floating platforms closely analogous to ours. Our platform is representative of small free-floating robots used in microgravity laboratories and shallow-water inspection tasks, where sensor-free fault adaptation is critical due to the absence of reliable fault-sensing hardware.

\section{Problem Formulation}
\label{sec:problem}

\subsection{Robot and Task}

We use a floating-platform robot controlled by eight independently-actuated thrusters plus one reaction wheel. Commands $\mathbf{u} \in [0,1]^9$ specify normalized thruster openings and wheel torque. The task is \emph{Go-to-Position}: navigate from a random spawn within a 3\,m radius to a fixed goal and hold position within 5\,cm for 50 consecutive steps.

The nominal observation vector $\otask \in \mathbb{R}^{15}$ contains relative goal position and orientation, body-frame velocity, and angular rate.

\subsection{Failure Model}

We model three independent failure modes acting on thruster $i$:
\begin{equation}
  u_i^\text{applied} = s_i \cdot u_i + \delta_i
  \label{eq:failure}
\end{equation}
where $s_i \in [0,1]$ is a scale factor and $\delta_i \in [0,1]$ is an additive offset. The three modes and their parameterizations are:
\begin{itemize}
  \item \textbf{DEG} (continuous degradation): $\delta_i = 0$, $s_i \sim \mathcal{U}(0, 1)$. Models biofouling or partial blockage.
  \item \textbf{DEAD} (abrupt failure): $\delta_i = 0$, $s_i = 0$. Models total actuator loss.
  \item \textbf{STK} (stuck-on): $s_i = 1$, $\delta_i \sim \mathcal{U}(0, 1)$. Models a jammed-open valve injecting constant thrust.
\end{itemize}

The degradation state is the 16-dimensional vector $\dgt = [s_0,\ldots,s_7,\,\delta_0,\ldots,\delta_7] \in [0,1]^{16}$. At each episode reset, exactly $k \sim \mathcal{U}\{0,\ldots,k_\text{max}\}$ thrusters are assigned a failure mode sampled from the per-mode probability vector $\mathbf{p}_\text{mode}$; remaining thrusters have $s_i=1$, $\delta_i=0$. A curriculum increases $k_\text{max}$ from 0 to 4 over training.

\subsection{Reinforcement Learning Formulation}

We model the task as a Markov Decision Process $(\mathcal{S}, \mathcal{A}, \mathcal{P}, r, \gamma)$, where $\mathcal{S}$ is the state space, $\mathcal{A}$ is the action space, $\mathcal{P}$ is the transition model, $r$ is the reward and $\gamma$ is the discount factor. The per-step reward is:
\begin{equation}
\begin{aligned}
  r_t = {} & w_p \,e^{-d_t/\alpha_p} + w_h \,e^{-|\theta_t|/\alpha_h} - w_{lv} \operatorname{clip}(v_t) \\
           & - w_{av} \operatorname{clip}(\omega_t) - w_b \,e^{-(d_{\max} - d_t)/\alpha_b}
\end{aligned}
\label{eq:reward}
\end{equation}
where $d_t{=}\|\mathbf{p}_t{-}\mathbf{p}^\star\|_2$ is position error, $\theta_t$ is heading error, $v_t$ and $\omega_t$ are linear and angular speed, and $d_{\max}$ is the workspace boundary radius. Weights: $w_p{=}1.0$, $w_h{=}0.25$, $w_{lv}{=}0.05$, $w_{av}{=}0.1$, $w_b{=}10.0$; scales $\alpha_p{=}\alpha_h{=}\alpha_b{=}1.0$.

An episode succeeds if $\|\mathbf{p}{-}\mathbf{p}^\star\|_2 < 0.05$\,m for at least 50 consecutive steps. Two metrics are reported across 5\,120 episodes per condition: the \emph{success rate} (SR, fraction of episodes meeting the criterion above) and the \emph{final position error} (FPE), the distance $\|\mathbf{p}_T{-}\mathbf{p}^\star\|_2$ at the last step $T$ of each successful episode (averaged over those episodes). SR captures whether the robot reaches the goal at all; FPE captures how precisely it settles once it does.

\section{Method: Privileged Actor-Critic for Fault Adaptation}
\label{sec:method}

\subsection{Asymmetric Actor-Critic Setup}
\begin{figure}[t]
\centering
\begin{tikzpicture}[
  font=\small,
  box/.style={draw, rounded corners=3pt, minimum width=1.8cm, minimum height=0.58cm, align=center},
  bigbox/.style={draw, rounded corners=4pt, inner sep=5pt},
  arr/.style={-{Stealth[length=4pt]}, thick},
  optarr/.style={-{Stealth[length=4pt]}, gray!70, thick, dashed},
  critarr/.style={-{Stealth[length=4pt]}, orange!80!black, thick},
]
\node[box, fill=gray!15]                       (otask) {$\otask\!\in\!\mathbb{R}^{15}$};
\node[box, fill=blue!25, right=1.3cm of otask] (pol)   {Policy\\$\pi_\theta$};
\node[box, fill=green!15, right=0.9cm of pol]  (act)   {$\mathbf{u}\!\in\![0,1]^9$};
\draw[arr] (otask) -- (pol);
\draw[arr] (pol)   -- (act);

\node[box, fill=gray!10, above=0.85cm of otask] (hist) {$\mathbf{H}\!\in\!\mathbb{R}^{32\!\times\!15}$};
\node[box, fill=blue!10] (obs) at (pol |- hist) {Observer $f_\phi$};
\draw[optarr] (hist) -- (obs);
\draw[optarr] (obs.south) -- node[right,font=\tiny,gray]{$\dhat$,\;$\bot$} (pol.north);
\node[box, draw=red!60!black, dashed, fill=red!5, right=0.7cm of obs] (mse) {MSE\\$\|\dhat{-}\dgt\|^2$};
\draw[optarr, red!70!black] (obs.east) -- (mse.west);

\node[box, draw=orange!80!black, fill=orange!10,
      below=0.85cm of pol] (crit) {Critic $V_\psi$\\{\scriptsize$[\otask\|\dgt]$}};
\node[box, draw=orange!70!black, fill=orange!5,
      left=1.3cm of crit]  (dgt_box)  {$\dgt\!\in\![0,1]^{16}$};
\draw[critarr] (dgt_box) -- (crit);
\draw[critarr] (otask.south) -- ++(0,-0.25) -| (crit.west);

\begin{scope}[on background layer]
  \node[bigbox, draw=gray!50, dashed,
    fit=(hist)(obs)(mse),
    label={[font=\tiny,gray!70,yshift=1pt]above:Observer extension (optional, \S\ref{subsec:obs})}] {};
  \node[bigbox, draw=orange!80!black, fill=orange!3,
    fit=(dgt_box)(crit),
    label={[font=\tiny,orange!80!black]below:\textit{training only}}] {};
\end{scope}
\end{tikzpicture}
\caption{Privileged actor-critic training setup. The asymmetric critic (orange, training only) receives $[\otask \,\|\, \dgt]$; its fault-informed value estimates shape policy gradients that encode fault-compensating behaviour into the actor without any fault sensing at deployment. \textbf{Base methods} (VAN-MLP-AC, GRU/LSTM-AC): actor receives only $\otask$. \textbf{Observer extension} (gray dashed, optional): actor additionally receives $\dhat$ from a history-based MLP, either OBS ($\lambda{=}0$, random projection) or OBS-MSE ($\lambda{=}1$, red dashed, supervised at 11\,pp cost).}
\label{fig:architecture}
\end{figure}
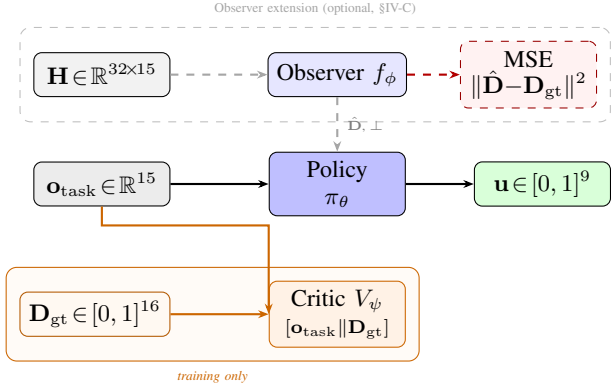

The core method is a PPO actor-critic in which the critic receives privileged access to the true degradation state $\dgt$ during training, while the actor receives only the standard task observation $\otask$ \cite{pinto2017asymmetric}. Figure~\ref{fig:architecture} shows the full setup.

\textbf{Actor.} At every step the actor receives $\otask$ and outputs a Gaussian action distribution:
\begin{equation}
  \pi_\theta(\mathbf{u} \mid \otask) = \mathcal{N}(\mu_\theta(\otask),\,\Sigma)
\end{equation}
No fault information reaches the actor at any point, neither during training nor at deployment. All fault-compensating structure must emerge from the policy gradient signal alone.

\textbf{Privileged critic.} The PPO critic receives the concatenated privileged observation $[\otask \,\|\, \dgt]$, including the ground-truth degradation state unavailable at deployment:
\begin{equation}
  V_\psi\bigl([\otask \,\|\, \dgt]\bigr) \in \mathbb{R}
\end{equation}
The critic's fault-aware value estimates produce lower-variance advantages that shape the policy gradient to encode fault-compensating behaviour. Because only the actor is executed at test time, exposing $\dgt$ to the critic carries no deployment cost: privileged fault access during training transfers to the actor as fault-tolerant policy structure, without the actor ever observing $\dgt$.

\subsection{Actor Architecture Variants}
We propose \textbf{RAFT} (Recurrent Asymmetric Fault Tolerant), a GRU-64 actor trained with the privileged critic above, and study four ablation variants under the same critic setup:
\begin{itemize}
  \item \textbf{VAN-MLP-AC}: feedforward MLP with hidden dimensions $[256, 128, 64]$. memory-less; the simplest ablation isolating the pure privileged-critic effect without recurrence.
  \item \textbf{GRU-256-AC} / \textbf{LSTM-64-AC} / \textbf{LSTM-256-AC}: Larger GRU and LSTM variants ablating the effect of hidden-state size and gating type.
\end{itemize}
All variants share the same privileged critic. The recurrent actors accumulate temporal failure signatures through their hidden state, without requiring an explicit history buffer.

\textbf{Reference policies.} Two non-RAFT policies frame the comparison: \textbf{VAN}, a standard MLP trained on a failure-free environment, serving as a failure-naive lower bound; and \textbf{Oracle}, the same GRU-64 architecture as RAFT but the actor additionally receives $\dgt$ at every step (concatenated to $\otask$). Oracle is trained on the same 3-mode failure curriculum as RAFT and serves as the per-mode upper bound, the SR achievable with perfect failure sensing at deployment.

\subsection{Observer Extension for Interpretability}
\label{subsec:obs}
We additionally study the \textbf{Observer (\methodname{})} architecture as an optional interpretability extension motivated by explicit system identification. It has no effect on the privileged critic; the critic architecture is identical to the base method.

\textbf{History buffer.} The most recent $L{=}32$ observations $\otask$ are stacked into $\mathbf{H} \in \mathbb{R}^{L \times 15}$, flattened to 480 features.

\textbf{Observer MLP.} A three-layer MLP maps the flattened history to a 16-dimensional latent:
\begin{equation}
  \dhat = f_\phi(\mathbf{H}) \in [0,1]^{16}
  \label{eq:observer}
\end{equation}
with hidden dimensions $[128, 64]$, ELU activations, and a sigmoid output. The augmented actor receives $\oaug = [\otask \,\|\, \dhat.{\tt detach}()]$; the detach operator prevents PPO gradients from updating $f_\phi$.

\textbf{OBS ($\lambda{=}0$).} In the default configuration $f_\phi$ receives no supervised gradient; its weights stay at random initialisation, acting as a fixed random projection of the history buffer. Experiment E6 confirms that OBS performs comparably to VAN-MLP-AC (64.4\,\% vs.\ 66.4\,\%, $p{>}0.05$), so the observer head's primary value is as a foundation for the OBS-MSE interpretability variant.

\textbf{OBS-MSE ($\lambda{=}1$).} Adding a supervised MSE loss:
\begin{equation}
  \mathcal{L} = \mathcal{L}_\text{PPO}(\theta) + \lambda\,\mathcal{L}_\text{mse}(\phi), \quad
  \mathcal{L}_\text{mse} = \|\dhat - \dgt\|_2^2
  \label{eq:combined}
\end{equation}
trains $f_\phi$ to estimate $\dgt$ from observable motion history. The observer output becomes human-readable, allowing a practitioner to inspect the policy's implicit fault belief at deployment, at a cost of 11\,pp SR (see E1 results).

\subsection{Training Protocol}
All policies are trained in Isaac Lab~\cite{mittal2023orbit} for 5\,000 PPO iterations using 4\,096 parallel environments and 24 steps per roll out ($\approx 5 \times 10^8$ environment steps total). The failure curriculum linearly increases $k_\text{max}$ from 0 to 4 over the first $5 \times 10^7$ steps. We use the Adam optimizer with a learning rate of $3 \times 10^{-4}$, 5 learning epochs per update, and 8 mini-batches. Three independent seeds (42, 7, 1337) are used; all results report mean $\pm$ std across seeds.
\section{Experiments and Results}
\label{sec:experiments}

We address four research questions through six experiments. All evaluations use 512 environments $\times$ 10 episodes per condition.
\textbf{E1, E3:} How close does RAFT approach the fault-oracle upper bound across failure count $k$ and failure mode?
\textbf{E2:} Is RAFT robust across the full continuous severity spectrum?
\textbf{E4:} Can RAFT adapt online to failures injected mid-episode without reconfiguration?
\textbf{E5, E6:} What is the contribution of recurrence and the asymmetric critic?

\subsection{E1 — Main Comparison (Mixed Modes)}

Five methods are compared under mixed-mode failures ($k \in \{0,\ldots,4\}$): VAN, Oracle, RAFT, OBS, and OBS-MSE (defined in Section~\ref{sec:method}). VAN-MLP-AC is reported separately in E6.

\begin{figure}[h]
  \centering
  \includegraphics[width=0.95\columnwidth]{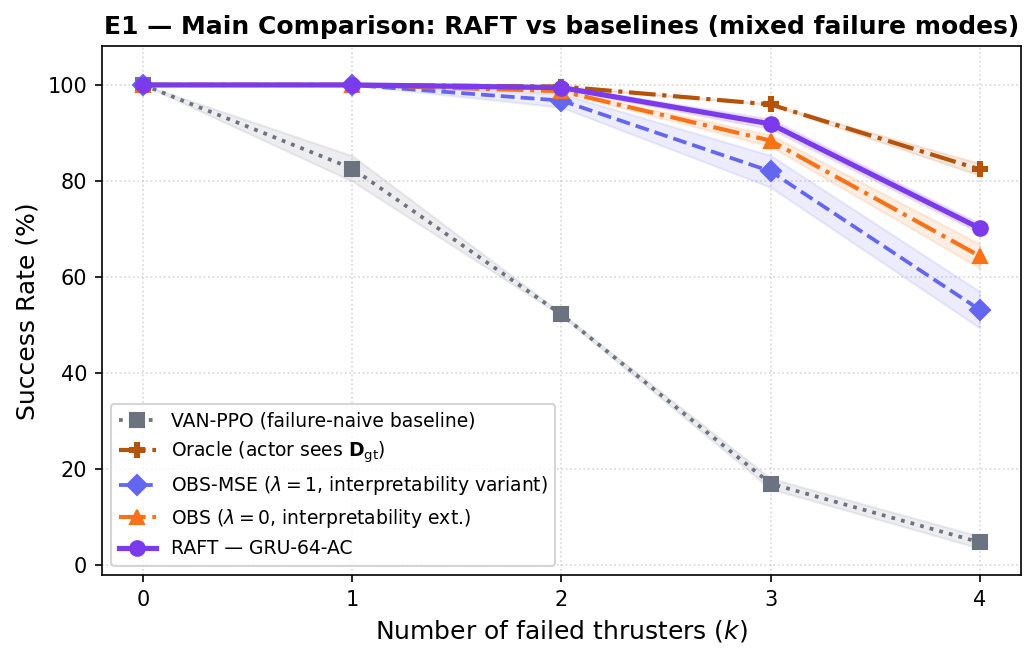}
  \caption{E1 — Success rate vs.\ failure count $k$ (mean\,$\pm$\,std, 3 seeds). RAFT (purple); OBS ($\lambda{=}0$, orange) and OBS-MSE (indigo): interpretability variants; Oracle: upper bound (actor sees $\dgt$); VAN: failure-naive baseline.}
  \label{fig:e1_sr}
\end{figure}

RAFT achieves \textbf{70.2\,\%} at $k{=}4$, closing \textbf{84\,\%} of the VAN$\to$oracle gap (4.8\,\%$\to$82.4\,\%), 12.2\,pp below the Oracle. All privileged-critic variants hold 100\,\% SR at $k{\leq}1$. VAN-MLP-AC (66.4\,\%, E6) shows the privileged critic alone closes 79\,\% of the gap; RAFT's recurrence adds a further 3.8\,pp SR on top. OBS (64.4\,\%) performs comparably to VAN-MLP-AC at no SR advantage. VAN collapses to 4.8\,\% at $k{=}4$, confirming that failure-curriculum training without privileged critic access is insufficient.

\subsection{E2 — Failure Severity Sweep}

With $k=1$ fixed, the severity parameter is swept across the full range for each mode independently: scale $\in \{0.0, 0.1, 0.3, 0.5, 0.7, 0.9, 1.0\}$ for DEG, offset $\in \{0.0, 0.1, 0.2, 0.4, 0.6, 0.8, 1.0\}$ for STK.

Both RAFT and OBS ($\lambda{=}0$) achieve 100\,\% SR across the full severity spectrum for both DEG and STK at $k{=}1$ (all 3 seeds). No severity cliff was observed for either method. Precision differs substantially: RAFT's final position error (FPE, the settling distance to the goal at episode end on successful episodes) is consistently roughly half that of OBS across all severity levels (Table~\ref{tab:e2}). The single exception is STK severity~1.0, where RAFT's mean rises to $0.107\pm0.058$\,cm due to a single-seed outlier (median below 0.08\,cm). GRU recurrence provides more precise failure compensation at any severity, even when SR is identical.

\begin{table}[h]
\caption{E2 — Final position error (cm) vs.\ severity at $k{=}1$ (mean, 3~seeds). SR\,=\,100\,\% for all methods at every level. DEG scale $s$: 0\,=\,dead, 1\,=\,nominal. STK offset $\delta$: 0\,=\,no effect, 1\,=\,fully stuck open.}
\label{tab:e2}
\centering\small
\setlength{\tabcolsep}{4pt}
\begin{tabular}{@{}S[table-format=1.1] r r @{\hspace{1.4em}} S[table-format=1.1] r r@{}}
\toprule
\multicolumn{3}{c}{DEG \ (thrust scale $s$)} &
\multicolumn{3}{c}{STK \ (stuck offset $\delta$)} \\
\cmidrule(r){1-3}\cmidrule(l){4-6}
{$s$} & {RAFT (cm)} & {OBS (cm)} & {$\delta$} & {RAFT (cm)} & {OBS (cm)} \\
\midrule
0.0 & 0.059\ci{0.007} & 0.114\ci{0.017} & 0.0 & 0.060\ci{0.006} & 0.114\ci{0.016} \\
0.1 & 0.058\ci{0.007} & 0.111\ci{0.017} & 0.1 & 0.057\ci{0.007} & 0.105\ci{0.014} \\
0.3 & 0.057\ci{0.006} & 0.107\ci{0.015} & 0.2 & 0.058\ci{0.010} & 0.099\ci{0.014} \\
0.5 & 0.055\ci{0.005} & 0.102\ci{0.016} & 0.4 & 0.054\ci{0.006} & 0.095\ci{0.012} \\
0.7 & 0.054\ci{0.005} & 0.097\ci{0.013} & 0.6 & 0.055\ci{0.006} & 0.099\ci{0.011} \\
0.9 & 0.054\ci{0.005} & 0.093\ci{0.009} & 0.8 & 0.069\ci{0.010} & 0.108\ci{0.013} \\
1.0 & 0.055\ci{0.005} & 0.094\ci{0.010} & 1.0 & 0.107\ci{0.058} & 0.129\ci{0.018} \\
\bottomrule
\end{tabular}
\end{table}

\subsection{E3 — Multi-Failure Scalability}

The failure mode is fixed per sub-experiment (DEG, DEAD, or STK) and $k$ is swept $0\ldots4$ for RAFT, OBS ($\lambda{=}0$), and the Oracle. All three policies share the same training distribution (3-mode mixed failure curriculum); Oracle differs only in that its actor receives the 16-dim $\dgt$ at every step, providing a per-mode oracle upper bound.

The mode hardness ranking is \textbf{DEG $<$ STK $<$ DEAD} for all three policies, including the oracle: binary total loss is hardest to compensate even with full failure-state knowledge, because the disturbance is maximally abrupt and the remaining thrusters carry the entire control authority. At $k{=}4$, Oracle reaches 98.8\,\% on DEG, 84.8\,\% on STK, and only 80.1\,\% on DEAD. RAFT consistently outperforms OBS across all modes and $k$ values: at $k{=}4$, RAFT leads OBS by $+3.0$\,pp on DEG (96.2\,\% vs.\ 93.2\,\%), $+7.9$\,pp on DEAD (52.2\,\% vs.\ 44.3\,\%), and $+5.4$\,pp on STK (72.7\,\% vs.\ 67.3\,\%). The RAFT-to-oracle gap is small on the easier modes (2.6\,pp on DEG, 12.1\,pp on STK) and widest on DEAD (27.9\,pp), where privileged failure information is most valuable; this is the principal cost of operating without fault sensing at deployment.

\begin{figure}[h]
  \centering
  \includegraphics[width=0.95\columnwidth]{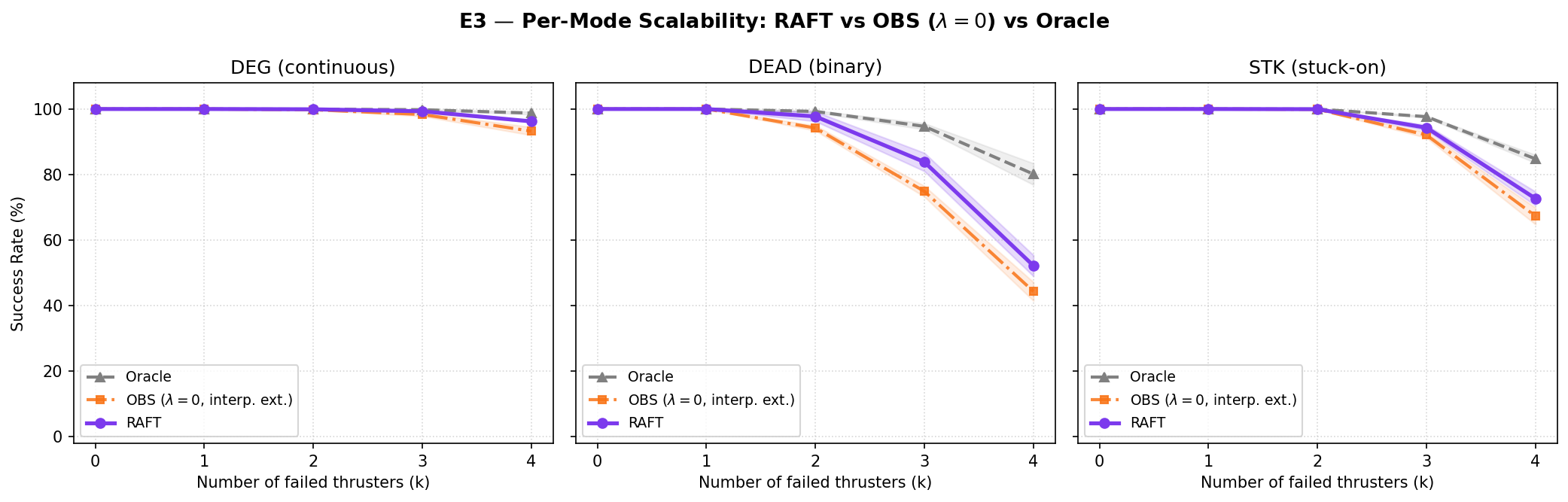}
  \caption{E3 — SR vs.\ $k$ per failure mode for RAFT (purple), OBS ($\lambda{=}0$, orange), and the Oracle (grey dashed). All three policies are evaluated under mode pinning. Mode hardness: DEG $<$ STK $<$ DEAD, holding even for the oracle.}
  \label{fig:e3_mode}
\end{figure}

\subsection{E4 — Mid-Episode Failure Injection}
\label{subsec:e4}

Episodes begin failure-free, and at step~100 of 400, $k$ thrusters are atomically degraded (no reset signal is given). Oracle receives the updated $\dgt$ instantly; RAFT relies on its GRU hidden state; VAN-MLP-AC uses a memory-less actor; OBS uses a sliding history buffer; VAN has no information.

The key finding: \textbf{VAN-MLP-AC's mid-episode drop nearly matches RAFT's}. VAN-MLP-AC drops $5.4$\,pp at $k{=}4$ ($61.0$\,\% vs reset-time $66.4$\,\%); RAFT drops $5.5$\,pp ($64.7$\,\% vs $70.2$\,\%). Because VAN-MLP-AC is memory-less, this parallel degradation confirms that the privileged critic's fault-aware value shaping, not GRU recurrence, is the primary mechanism for online adaptation. RAFT's GRU adds $3.7$\,pp on top, narrowing the Oracle gap to $10.8$\,pp at $k{=}4$ (vs $14.5$\,pp for VAN-MLP-AC and $26.3$\,pp for OBS). VAN collapses immediately to $6.8$\,\% at $k{=}4$. OBS's larger $15.2$\,pp drop reflects its sliding history buffer needing more steps to accumulate a new failure signature; the Oracle itself drops only $6.9$\,pp from its reset-time SR ($82.4$\,\%\,$\to$\,$75.5$\,\%), establishing that mid-episode adaptation is intrinsically harder than reset-time even with perfect failure information.

\begin{figure}[h]
  \centering
  \includegraphics[width=0.95\columnwidth]{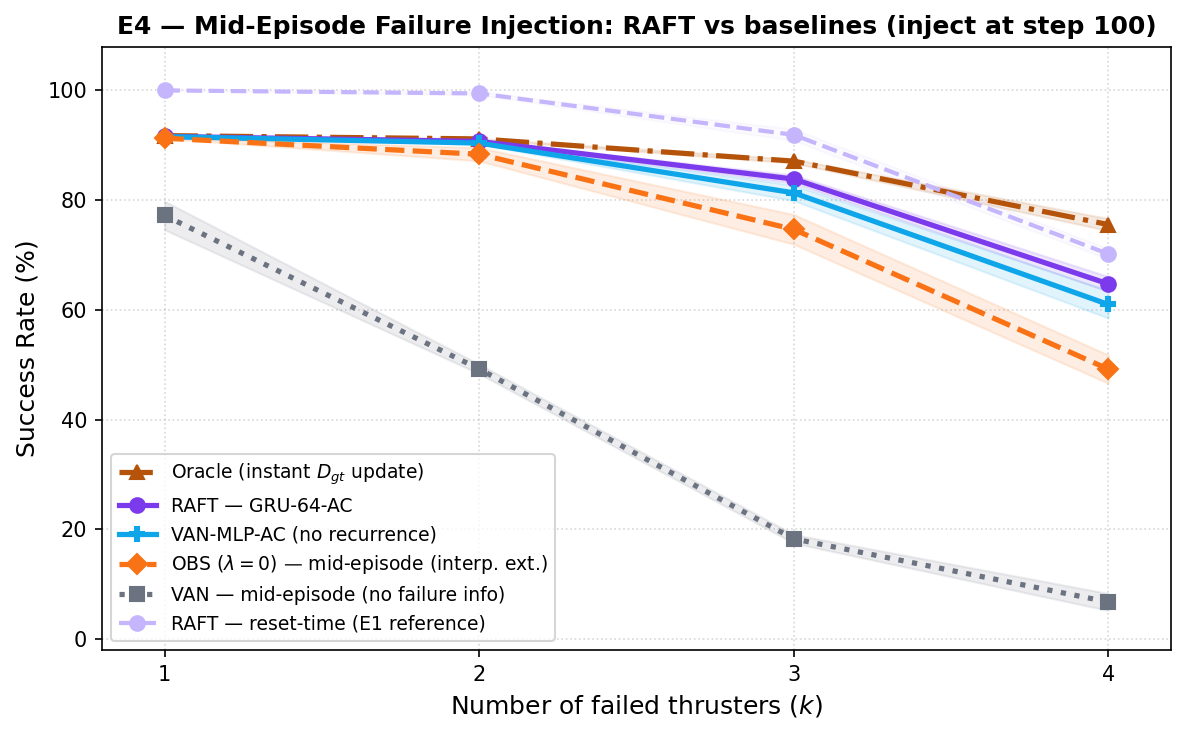}
  \caption{E4 — Mid-episode failure injection at step~100. VAN-MLP-AC (sky blue) tracks RAFT (purple) closely, confirming the privileged critic, not recurrence, drives adaptation. Dashed light-purple: RAFT reset-time SR (E1) reference (5.5\,pp drop at $k{=}4$). OBS (orange) lags 15.5\,pp due to its slower history buffer.}
  \label{fig:e4}
\end{figure}

\subsection{E5 — Recurrent Policy Ablation}
\label{subsec:e5}

Four recurrent policies (GRU and LSTM, hidden dims 64 and 256) are trained for 5\,000 iterations on the same failure curriculum as RAFT, receiving only instantaneous $\otask$ (no asymmetric critic, no history buffer). This tests whether step-by-step recurrence alone can implicitly detect failures from the policy gradient signal.

All four variants track VAN-MLP closely and collapse at high $k$ (at $k{=}4$: GRU-64 4.0\,\%, GRU-256 1.9\,\%, LSTM-64 3.3\,\%, LSTM-256 3.0\,\%, vs.\ VAN-MLP 4.8\,\%). Wide variance ($>13$\,pp at $k{=}1$) indicates seed-dependent local optima rather than reliable failure detection. Recurrence alone cannot achieve fault tolerance: the 65-pp gap to RAFT (70.2\,\%) is attributable entirely to the absent asymmetric critic, which E6 confirms by recovering 66.4\,\% with a memory-less actor under the same critic.

\begin{figure}[h]
  \centering
  \includegraphics[width=0.95\columnwidth]{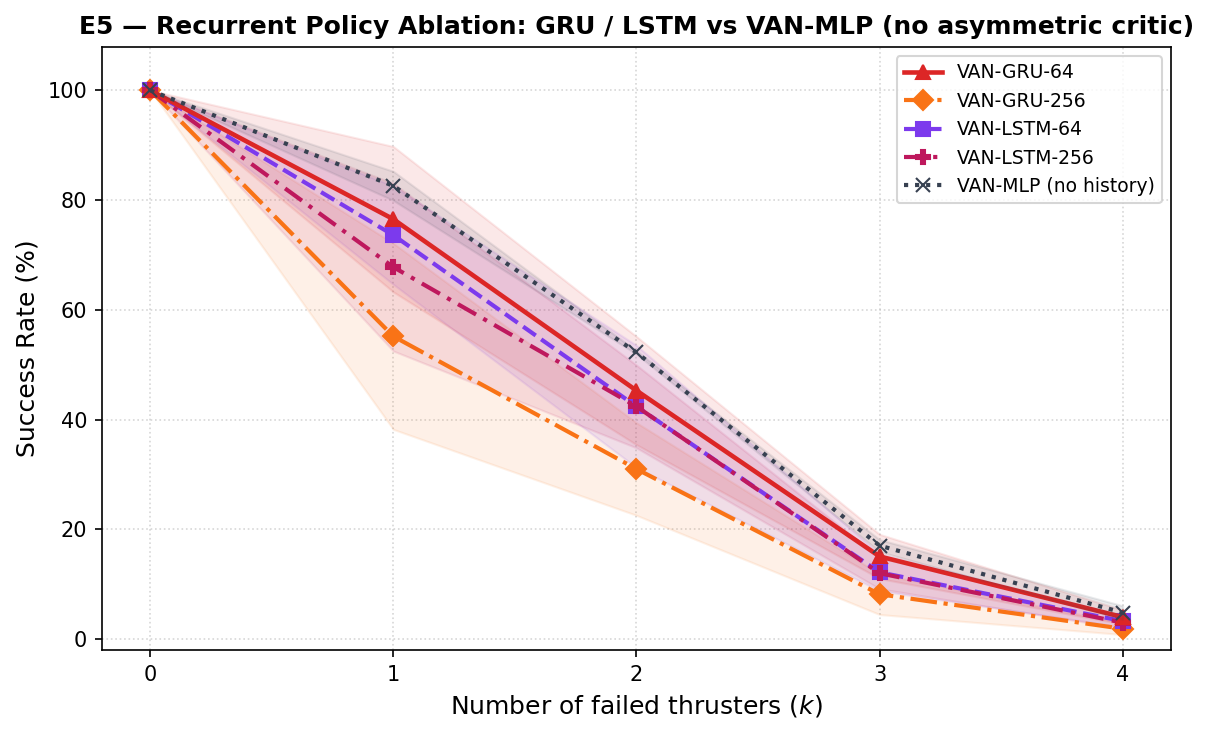}
  \caption{E5 — Recurrent baselines (GRU/LSTM, hidden dims 64/256) trained without the asymmetric critic all track VAN-MLP and collapse at high $k$. Wide variance ($>13$\,pp at $k{=}1$) indicates unstable local optima. Recurrence alone cannot substitute for privileged critic access.}
  \label{fig:e5}
\end{figure}

\subsection{E6 — Asymmetric Critic Ablation}
\label{subsec:e6}

To isolate the asymmetric critic's contribution from the observer head and history buffer, five variants (VAN-MLP-AC, RAFT/GRU-64-AC, GRU-256-AC, LSTM-64-AC, LSTM-256-AC) are trained with $\dgt$ available to the critic but \emph{not} to the actor. No observer head or history buffer is used. RAFT and VAN-MLP (no AC, from E1) serve as upper and lower references.

The striking finding: \textbf{VAN-MLP-AC} (memory-less, no history, no observer) achieves \textbf{66.4\,$\pm$\,2.9\,\%} at $k{=}4$, a 61-pp jump over the same architecture without the asymmetric critic (VAN-MLP at 4.8\,\%, E1). The asymmetric critic alone is sufficient to recover most of the fault-tolerance signal: the value function's privileged access to $\dgt$ during training shapes policy gradients that generalize to failure scenarios at inference, without any explicit failure state in the actor.

\begin{figure}[h]
  \centering
  \includegraphics[width=0.95\columnwidth]{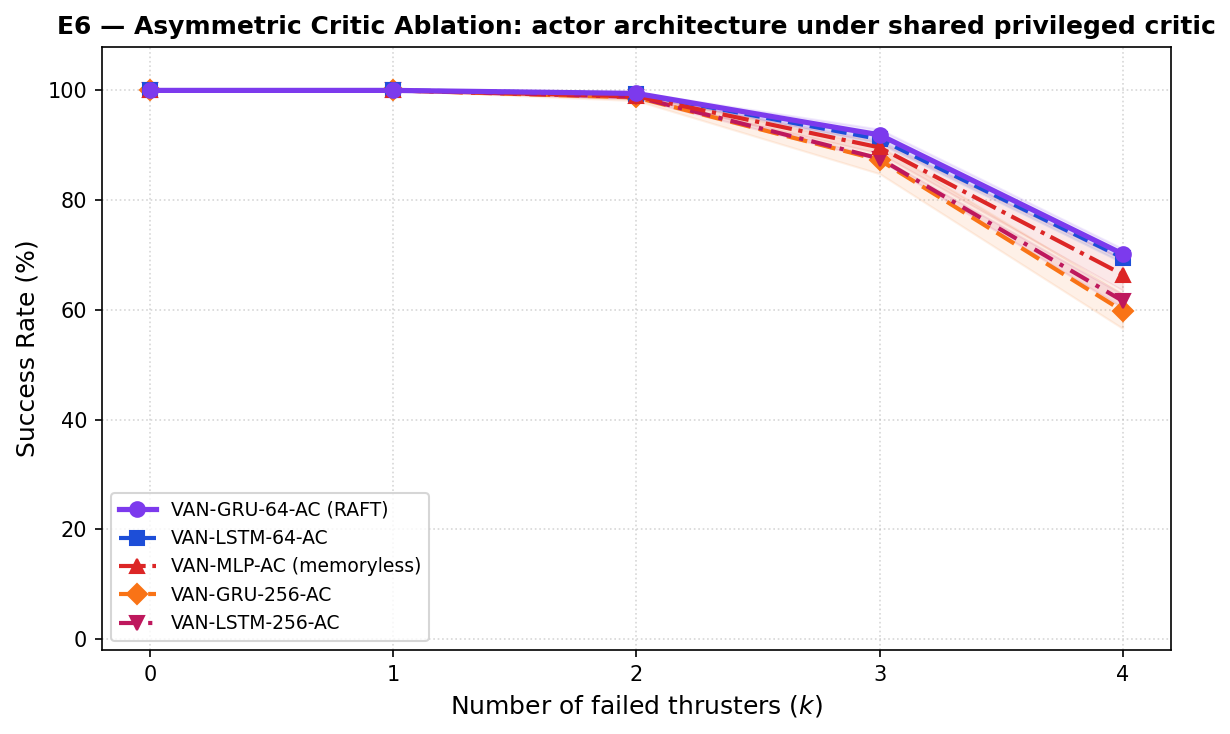}
  \caption{E6 — Five AC variants share the same privileged critic; the actor differs only in architecture. VAN-MLP-AC (66.4\,$\pm$\,2.9\,\%) already recovers most of the gap; compact recurrent variants RAFT/GRU-64-AC (70.2\,$\pm$\,1.0\,\%) and LSTM-64-AC (69.5\,$\pm$\,0.9\,\%) add 3--4\,pp; larger hidden dims (256) underperform dim-64 counterparts.}
  \label{fig:e6}
\end{figure}

Compact recurrent variants RAFT/GRU-64-AC (70.2\,$\pm$\,1.0\,\%) and LSTM-64-AC (69.5\,$\pm$\,0.9\,\%) exceed VAN-MLP-AC by 3--4\,pp, plausibly through richer temporal context. Larger hidden dims (256) underperform their dim-64 counterparts in both families, consistent with E5. All AC variants vastly outperform E5's best recurrent baseline (4.0\,\%), confirming the 65-pp gap was driven by the absent critic, not by actor architecture choices.

\section{Discussion}
\label{sec:discussion}

\textbf{Why does privileged critic training enable sensor-free fault adaptation?} The value function's access to $\dgt$ produces fault-aware advantage estimates: episodes where the policy fails to compensate for a specific failure pattern receive lower advantage, penalizing fault-naive behavior in the gradient. The critic acts as a fault-informed teacher that shapes the actor through the gradient signal rather than through direct observation. E5 confirms this mechanism: removing the privileged critic collapses performance to VAN-level (best recurrent: 4.0\,\%), while restoring it (E6) recovers 66.4\,\% for a memory-less actor.

\textbf{Recurrence adds modest gains on top of the critic.} Without privileged critic access, recurrence fails (best: 4.0\,\% at $k=4$). With it, compact recurrent actors (RAFT: 70.2\,\%, LSTM-64-AC: 69.5\,\%) edge out memory-less VAN-MLP-AC (66.4\,\%) by 3--4\,pp; larger hidden states (dim 256) hurt in both GRU and LSTM families, consistent with over-parameterization on this task.

\textbf{Interpretability has a quantified cost.} The MSE objective imposes a representation constraint: the 16-dim observer output must align with $\dgt$, introducing a bottleneck wherever fault-adaptive control needs features not well-described by $\dgt$ directly. The result is a clear practitioner choice: RAFT for maximum performance, OBS for a competitive feedforward option, or OBS-MSE for a human-readable fault estimate at the cost of 11\,pp SR. Task-motivated observer objectives (contrastive, reward-supervised) may recover OBS-MSE's interpretability without the performance cost.

\textbf{Mid-episode adaptation (E4).} The near-identical mid-episode drops for memory-less VAN-MLP-AC (5.4\,pp) and RAFT (5.5\,pp) confirm that the privileged critic's value shaping, not recurrence, is the dominant adaptation mechanism. RAFT's GRU contributes a further 3.7\,pp through richer temporal context; OBS's 15.2\,pp drop reflects its sliding buffer needing more steps to accumulate a new failure signature.

\textbf{Where is fault tolerance encoded?} RAFT's actor never observes $\dgt$, yet compensates for failures at deployment. The mechanism is not runtime fault estimation: E5 shows that recurrence alone (which would in principle accumulate failure signatures into the hidden state) achieves only $\le 4\,\%$ SR at $k{=}4$ without the privileged critic. Instead, the gradient-shaped policy weights themselves encode fault-compensating behavior, and the GRU contributes temporal context for richer actuation rather than acting as a fault detector. A separate question is the per-mode observability of $\dgt$ itself: DEG and DEAD induce visible changes in motion dynamics, but a stuck-open offset (STK) produces a constant thrust injection independent of the command and is plausibly unobservable from passive motion history alone. Closing such observability gaps likely requires active-excitation strategies.

\textbf{Limitations.} Experiments are conducted entirely in simulation; sim-to-real transfer is not evaluated. The task is 2D navigation; 6-DOF tasks with coupled pitch/roll failures may require longer history or recurrent observers. The Transformer history-encoder baseline (HIS) did not converge in our setting.

\section{Conclusion}
\label{sec:conclusion}

Privileged critic training is sufficient for sensor-free thruster fault adaptation: RAFT achieves \textbf{70.2\,\%} SR at $k{=}4$, closing 84\,\% of the gap from a failure-naive baseline (4.8\,\%) to the Oracle (82.4\,\%) while its actor never observes $\dgt$ at deployment. VAN-MLP-AC (memory-less, same critic) closes 79\,\% of the gap, isolating the critic as the primary mechanism; RAFT's GRU adds a further 3.8\,pp SR. Recurrent policies without privileged critic access reach at most 4.0\,\%, confirming recurrence alone cannot substitute.

The Observer extension adds interpretability without performance: OBS matches VAN-MLP-AC (64.4\,\%), and OBS-MSE makes the fault estimate human-readable at a cost of 11\,pp SR, a quantified tradeoff for practitioners who require explainable estimates. The near-identical mid-episode drops for memory-less VAN-MLP-AC and recurrent RAFT (5.4 vs.\ 5.5\,pp) suggest fault tolerance is encoded in the gradient-shaped policy weights rather than in a runtime state estimator.

Future work will extend to 6-DOF platforms, evaluate sim-to-real transfer, and investigate task-motivated observer objectives to close the STK observability gap.

\bibliographystyle{IEEEtran}
\bibliography{references}

\end{document}